\documentclass{article}
\usepackage{graphicx} 
\usepackage{tabularx}
\usepackage{array}
\usepackage{authblk}
\usepackage{booktabs}
\usepackage[margin=1in]{geometry}
\usepackage{xcolor}
\usepackage{amssymb}
\title{GTO Wizard Benchmark}
\author{Marc-Antoine Provost}
\author{Nejc Ilenic}
\author{Christopher Solinas}
\author{Philippe Beardsell}
\affil{\{marco,nejc,chris,phil\}@gtowizard.com \\ GTO Wizard}
\date{March 2026}

\begin{document}

\maketitle

\begin{abstract}
We introduce GTO Wizard Benchmark, a public API and standardized evaluation framework for benchmarking algorithms in Heads-Up No-Limit Texas Hold'em (HUNL).
The benchmark evaluates agents against GTO Wizard AI, a state-of-the-art superhuman poker agent that approximates Nash Equilibria, and defeated Slumbot, the 2018 Annual Computer Poker Competition champion and previous strongest publicly accessible HUNL benchmark, by $19.4$ $\pm$ $4.1$  bb/100.
Variance is a fundamental challenge in poker evaluation; we address this by integrating AIVAT \cite{aivat}, a provably unbiased variance reduction technique that achieves equivalent statistical significance with ten times fewer hands than naive Monte Carlo evaluation.
We conduct a comprehensive benchmarking study of state-of-the-art large language models under zero-shot conditions, including GPT-5.4, Claude Opus 4.6, Gemini 3.1 Pro, Grok 4, and others.
Initial results and analysis reveal dramatic progress in LLM reasoning over recent years, yet all models remain far below the baseline established by our benchmark.
Qualitative analysis reveals clear opportunities for improvement, including representation and the ability to reason over hidden states.
This benchmark provides researchers with a precise and quantifiable setting to evaluate advances in planning and reasoning in multi-agent systems with partial observability.
\end{abstract}

\section{Introduction}
Games have long been an integral part of the field of artificial intelligence, by providing difficult but easily verifiable benchmarks that test similar sets of skills that we would expect from strong agents -- reasoning, strategic planning, and sequential decision-making. More recently, advances in Large Language Models (LLMs) have demonstrated remarkable capabilities, surpassing traditional methods across diverse domains, including code generation and mathematical problem-solving. As benchmarks created by humans become increasingly saturated, a natural testing ground for assessing the ever-growing capabilities of these models is games. Poker stands out as a quintessential example of a game that requires this multifaceted skill set. Success in poker depends on a player's ability to estimate their opponent(s)' holdings by incorporating all the information available to them -- actions, play style, and the overall game context. This requires a sophisticated blend of mathematical calculation, reasoning, memory, long-term planning, and game theory.

Despite the growing interest in applying generalist AI agents and LLMs to games, a standardized platform for benchmarking their performance in poker has been lacking. Human-versus-agent matches have long been a standard means for evaluation (\cite{deepstack, libratus}), but they are costly and challenging to organize---meaning they are feasible only for research initiatives with significant resources. We introduce GTO Wizard Benchmark: a public API that provides a standardized environment for benchmarking agents in Heads-Up No-Limit Texas Hold'em. Our benchmark is designed not as a static evaluation platform, but as an evolving ecosystem for poker AI research. Importantly, though this initial release only supports the Heads-Up No-Limit Texas Hold'em format, our benchmark evaluates against a general poker agent capable of playing a variety of formats. Rather than building an isolated, highly optimized Heads-Up agent, we developed a general agent able to play a wide variety of two-player and multi-player scenarios, including varying stack sizes, cash game formats (rake, antes, etc.), and tournament configurations. Due to this foundation, we have high ambitions for the future of the platform, including continually improving our agents, expanding to other game variants such as Pot-Limit Omaha, and introducing support for more than two players.

GTO Wizard Benchmark offers several key contributions:
\begin{itemize}
    \item Our API enables researchers to test their agents against GTO Wizard AI. This proprietary, state-of-the-art poker agent demonstrated superior performance against Slumbot, the past winner of the Annual Computer Poker Competition (ACPC)\cite{acpc}.
    \item Our system evaluates agents using AIVAT\cite{aivat}, a provably unbiased variance reduction technique for assessing performance in imperfect information games, which allows agents to achieve the same statistical significance with ten times less data.
    \item We foster collaboration and transparency by making all evaluation results publicly available through a real-time leaderboard.
    \item We perform an extensive evaluation of state-of-the-art general-purpose LLMs against our superhuman baseline. We find that the best-performing model, GPT-5.3 (Extra High reasoning), achieves a luck-adjusted win rate of -16 $\pm$ 3.0 bb/100 (big blinds per hundred hands), demonstrating a massive leap in reasoning capabilities compared to previous-generation models.
\end{itemize}

We provide code examples for researchers to get started with benchmarking their agents.\footnote{https://github.com/gtowizard-ai/researcher-api-client}

\section{Related Work}
For decades, games have been a standard testbed for evaluating AI models. Hence, the development of superhuman-level agents in games has a long history \cite{campbell2002deep, silver2016mastering}. Significant progress was made in developing superhuman AI agents in poker between 2015 and 2020, driven by algorithmic advances utilizing neural networks and depth-limited solving \cite{deepstack, libratus, 2017Sci...356..508M, brown2019superhuman, supremus, rebel}. While the creation of such agents accelerated during the mid-to-late 2010s, evaluating these agents remained painstakingly slow, with the primary means of evaluation being human-versus-agent or agent-versus-agent matches. Human-versus-agent matches are extremely hard to setup; top-tier professionals have high opportunity costs and the compensation offered need to match the expected value these professionals can generate by playing high-stakes games. Furthermore, because of the variance inherent to poker, these matches have to span multiple weeks in order to get statistical significance \cite{libratus}.
The resources involved in coordinating this type of evaluation prohibit smaller research labs from measuring progress meaningfully, which slows the development of new algorithms and methods.

On the other hand, while agent head-to-head performance was prominent thanks to the ACPC, the competition was discontinued in 2018---around the same time when the first superhuman agents in No-Limit Hold'em were developed. The competition's strict computational requirements was no longer appropriate for the new reasoning paradigm. In 2021, Eric Jackson, the creator of Slumbot\cite{jackson2013slumbot}, an abstraction-based agent and the last winner of the ACPC, released a public API\footnote{https://slumbot.com/}, enabling researchers and hobbyists to benchmark their poker agent against it. While useful, its abstraction-based method limits its action space, leaving it vulnerable to potential exploits and making it less rigorous for scientific benchmarking. Moreover, the lack of a strong variance reduction technique for evaluation makes it expensive to obtain statistically significant results. Recently, Zhuang et al. developed PokerBench\cite{zhuang2025pokerbench}, a poker benchmark consisting of 11,000 solved poker scenarios and decision points for evaluating LLMs' poker-playing abilities. While substantial, this benchmark consists of handpicked human situations and does not encompass the full spectrum of poker scenarios needed to evaluate AI agents adequately. GTO Wizard Benchmark enables scientifically rigorous benchmarking of AI agents in poker.

\section{GTO Wizard Benchmark}
GTO Wizard\footnote{https://gtowizard.com/} is the premier learning platform for poker enthusiasts, offering a suite of tools to learn and study poker theory. Central to this platform is GTO Wizard AI, a proprietary poker agent that combines equilibrium-finding algorithms with deep learning.
Unlike static agents that rely on abstractions and precompute their strategy ahead of time, such as Slumbot, GTO Wizard AI computes strategies in real-time. Trained through self-play reinforcement learning over hundreds of millions of hands, it approximates Nash Equilibrium strategies with high fidelity across arbitrary stack depths and bet sizes. In 2022, GTO Wizard AI demonstrated its superiority by defeating Slumbot, the 2018 ACPC champion, by a margin of 19.4 $\pm$ 4.1 bb/100 over 150 000 hands. As a reference point, strong professional poker players have a win rate that oscillates around 5 bb/100. This result currently establishes GTO Wizard AI as the strongest poker agent in No-Limit Texas Hold’em and validates its usefulness for benchmarking purposes.
We expose this agent as a standardized opponent through our publicly available API. The environment is a Heads-Up No-Limit Texas Hold’em (HUNL) match with the following characteristics:
\begin{itemize}
\item It evaluates agents with the standardized set of rules used in the ACPC. The blinds are 50 and 100, with the player stack sizes at 200 big blinds, i.e., 20 000 chips.
\item It is implemented as a RESTful API that handles game state management, reducing the engineering overhead for researchers.
\item It integrates AIVAT \cite{aivat}, an unbiased variance-reduction method. Due to the high variance inherent to poker, a winning player could lose over a relatively large sample of hands due to bad card distribution and player action randomness. AIVAT achieves a threefold reduction in standard deviation, resulting in the same statistical significance with 10 times less data.  
\end{itemize}

\section{Experiments}
To evaluate the capabilities of current state-of-the-art LLMs in HUNL, we benchmarked them against GTO Wizard AI. Their luck-adjusted bb/100 is used to evaluate their performance in HUNL. Note that this corresponds to their AIVAT score, expressed in bb/100. Below are some of the evaluation metrics returned by GTO Wizard Benchmark's API:
\begin{itemize}
    \item AIVAT score: how many chips the agent was expected to win, adjusted for luck.
    \item Chips: how many absolute chips were won or lost by the agent.
    \item Total number of hands played.
\end{itemize}
We report the full list of evaluation metrics returned by GTO Wizard Benchmark’s API in the Appendix.

\subsection{Setup}
We used our GTO Wizard Benchmark API to play 5000 poker hands of HUNL Texas Hold'em per model. Blinds were 50 and 100 chips, with stack sizes of 200 big blinds. In this formulation, each hand is treated as an independent event, with player chip stacks resetting to 20 000 before each hand begins. This follows the tradition of the ACPC.

We prompted each model with a text-based representation of the game history and instructed it to output a valid poker action alongside the betting amount if the action was a bet or a raise. We also instructed the models to provide their reasoning behind each action. The exact prompts used are included in the Appendix. No external tools or equity calculators were provided; the agents relied solely on their internal reasoning capabilities in a zero-shot setting. We evaluated the following LLMs: GPT-5.4 with Extra High reasoning, GPT-5.4 Mini and Nano, GPT-5.3 Extra High and High reasoning, Claude Opus 4.6, Claude Opus 4.5, Gemini 3.1 Pro, Gemini 3 Pro, Gemini 2.5 Pro, Grok 4 High reasoning, Kimi K2.5, GPT-4o, and GPT-4. We also evaluated a few baseline agents: 
\begin{itemize}
    \item Check Call Agent: An agent who always checks if he has the possibility. Otherwise, it calls the opponent's bet.
    \item Always Fold Agent: An agent that always folds or checks if he can't fold.
    \item Uniform Random Agent: An agent that takes actions with a uniform random policy.
    \item All-In Agent: An agent that always goes all-in.
\end{itemize}
We report the cost and total runtime of each model for our experiments in \ref{tab:agent_cost_runtime}.

\begin{table}[h]
    \centering
    \begin{tabular}{lcc}
        \toprule
        \textbf{Model} & \textbf{Total Cost (USD)} & \textbf{Total Runtime (hours)} \\
        \midrule
        GPT-5.4 Extra High reasoning & 1915.02 & 635.93 \\
        GPT-5.3 Extra High reasoning & 1041.19 & 446.36\\
        Grok 4 High reasoning & 612.09 & 314.53\\
        GPT-4 & 403.2 & 12.94 \\
        Claude Opus 4.6 & 323.07 & 55.16\\
        GPT-5.3 High reasoning & 287.82 & 127.2 \\
        Gemini 3 Pro & 176.98 & 47.52 \\
        Claude Opus 4.5 & 141.56 & 13.96 \\
        Gemini 2.5 Pro & 122.52 & 35.27 \\
        Gemini 3.1 Pro & 72.22 & 19.86\\
        Kimi K2.5 & 59 &  206.38\\
        GPT-4o & 34.49 & 7.12\\
        GPT-5.4 Mini & 15.29 & 4.42\\
        GPT-5.4 Nano & 3.27 & 3.32\\
        \midrule
        \textbf{Total} & \textbf{5207.72} & \textbf{1929.98} \\ 
        \bottomrule
    \end{tabular}
    \caption{Cost and runtime for LLM agent evaluation}
    \label{tab:agent_cost_runtime}
\end{table}

\subsection{Results}
We report the results for agents and then provide some analysis and insights into model performance. We summarize their performance in Table \ref{tab:luck_adjusted_win_rate}.

\begin{table}[ht]
    \centering
    \begin{tabular}{lc}
        \toprule
        \textbf{Agent} & \textbf{Luck-adjusted win rate (bb/100)} \\
        \midrule
        \textbf{GPT-5.3 Extra High reasoning} & \textbf{-16 $\pm$ 3} \\
        GPT-5.4 Extra High reasoning & -17.8 $\pm$ 3.7\\
        GPT-5.3 High reasoning & -18.2 $\pm$ 3.9 \\
        Claude Opus 4.6 & -20.4 $\pm$ 8.6 \\
        Claude Opus 4.5 & -22.3 $\pm$ 10.1 \\
        Gemini 3.1 Pro & -30.8 $\pm$ 4.5 \\
        Gemini 3 Pro & -30.1 $\pm$ 7.5 \\
        Gemini 2.5 Pro & -39.2 $\pm$ 9.6 \\
        Kimi K2.5 & -41.4 $\pm$ 12.5 \\
        Always Fold & -64.6 $\pm$ 3.3 \\
        Grok 4 High reasoning & -60 $\pm$ 15.1 \\
        GPT-4o & -64.9 $\pm$ 15 \\
        GPT-5.4 Mini (No reasoning) & -107.9 $\pm$ 9.1\\
        GPT-4 &-136.2 $\pm$ 25.6  \\
        GPT-5.4 Nano & -189.7 $\pm$ 15.3 \\
        Check call & -241.1 $\pm$ 26.2\\
        Uniform Random & -284.9 $\pm$ 30 \\
        All-in & -380.6 $\pm$ 4.3 \\
        \bottomrule
    \end{tabular}
    \caption{Luck-adjusted win rate of evaluated agents in big blinds per hundred hands (bb/100). The $\pm$ term corresponds to the standard deviation of the match.}
    \label{tab:luck_adjusted_win_rate}
\end{table}
Across the board, these models struggle to capture the details and subtleties of poker strategy. The best performing agent is GPT-5.3 Extra High reasoning with a luck-adjusted win rate of -16 bb/100.
We note a steep improvement in reasoning models over the years, especially from the GPT model family. Over the span of 2 years and 9 months, their models improved from a luck-adjusted bb/100 of -136.2 to -16. 

We report the detailed performance metrics in the Appendix.
\subsection{Analysis}
In this section, we conduct a short analysis of model play and strategies compared to the strategy provided by GTO Wizard AI.

Upon investigation of its internal reasoning, the best performing model, GPT-5.3 Extra High reasoning, mismatches around 2\% of its holdings. That is, the model will sometimes mistake offsuit hands for suited ones and vice versa. This behavior mostly occurs preflop, when the agent is dealt his cards, and the error propagates throughout the hand. While this happens relatively infrequently, we observe this across all the LLMs evaluated. With earlier models, such as GPT-4, we can observe that the agent sometimes completely mistakes his own holdings. Some examples are shown in the Appendix.
These types of \textit{representational} errors can lead to serious mistakes in the agents' policies, and likely explain some, but not all, of the performance gap between the strategies of these agents and GTO Wizard AI.

To investigate the soundness of LLM agent gameplay, we visualize their preflop strategies. We depict the usual 13x13 combination of hands a player can hold, known as the range matrix. In these depictions, red represents the raise action, green represents the action of calling or checking, and blue indicates folding. In Figure 1, 2, and 3, we show the opening ranges (the first decision node in the game tree) of a few different models, namely GPT-5.3 Extra High reasoning, Claude Opus 4.6, and GTO Wizard AI. Compared to GTO Wizard's AI (near) optimal policy, we observe that GPT-5.3 is unbalanced and doesn’t raise or fold with the correct frequencies. In Figure 4, 5, and 6, we depict the Big Blind defending range against a bet from the Small Blind for GPT-5.3 Extra High reasoning, Claude Opus 4.6 and GTO Wizard AI. In this situation, GPT-5.3 only 3-bets the very best opening hands and calls with almost all of the rest.
This means that when GPT-5.3 3-bets, a clever opponent can easily deduce that it holds one of the strongest opening hands and respond accordingly.

\begin{figure}
    \centering
    \includegraphics[width=0.7\linewidth]{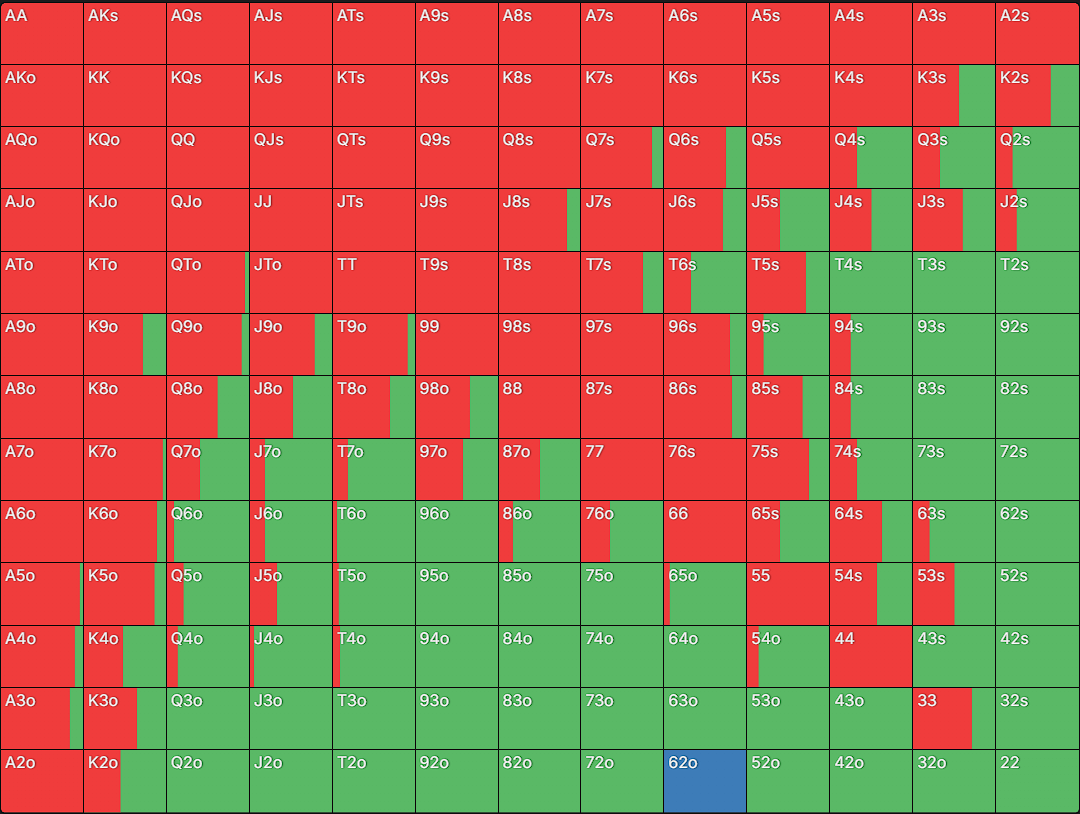}
    \caption{SB Open - GPT-5.3 Extra High reasoning}
    \label{fig:sb-open-gpt}
\end{figure}
\begin{figure}
    \centering
    \includegraphics[width=0.7\linewidth]{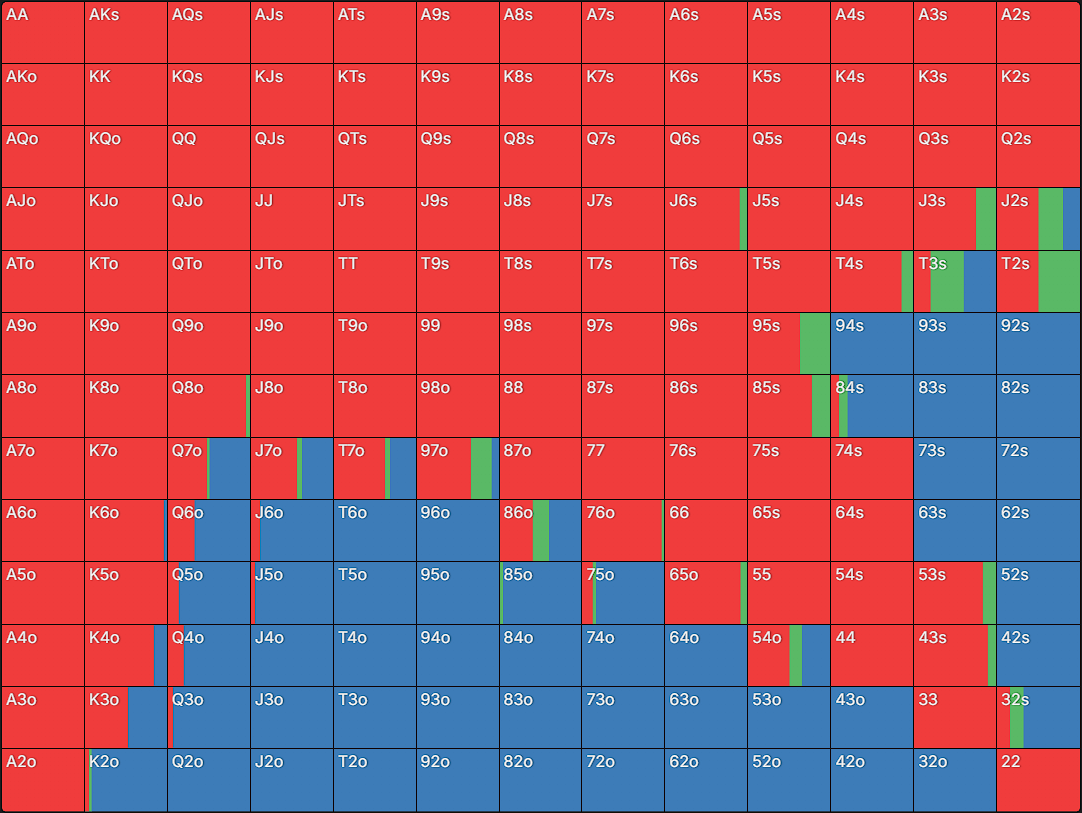}
    \caption{SB Open - Claude Opus 4.6}
    \label{fig:sb-open-claude}
\end{figure}
\begin{figure}
    \centering
    \includegraphics[width=0.7\linewidth]{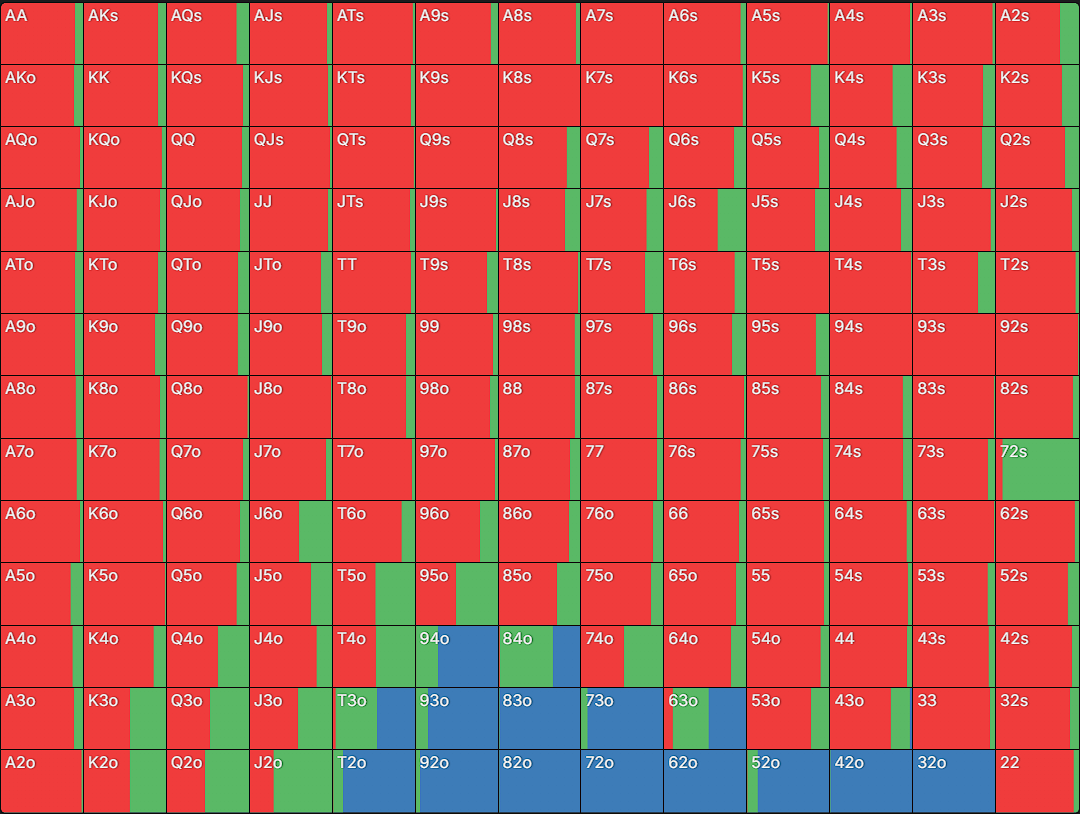}
    \caption{SB Open - GTO Wizard AI}
    \label{fig:sb-open-gtow}
\end{figure}
\begin{figure}
    \centering
    \includegraphics[width=0.7\linewidth]{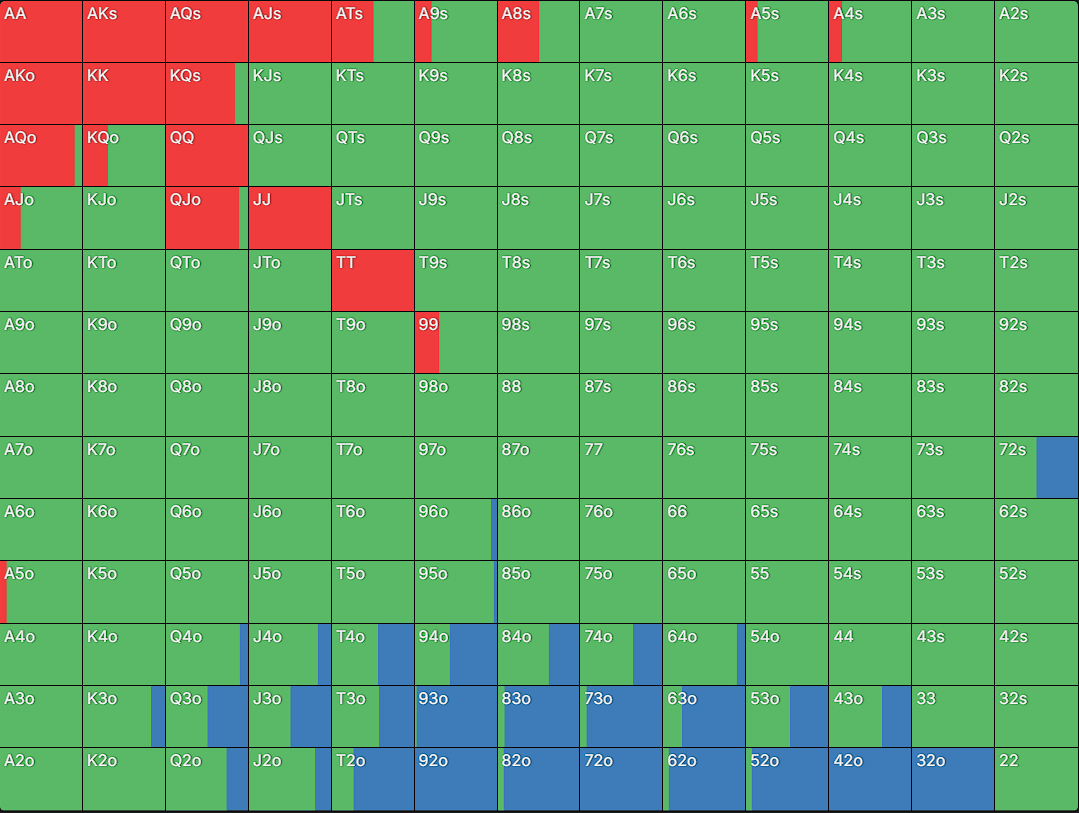}
    \caption{BB vs SB Open - GPT-5.3 Extra High reasoning}
    \label{fig:bb-vs-sb-gpt}
\end{figure}
\begin{figure}
    \centering
    \includegraphics[width=0.7\linewidth]{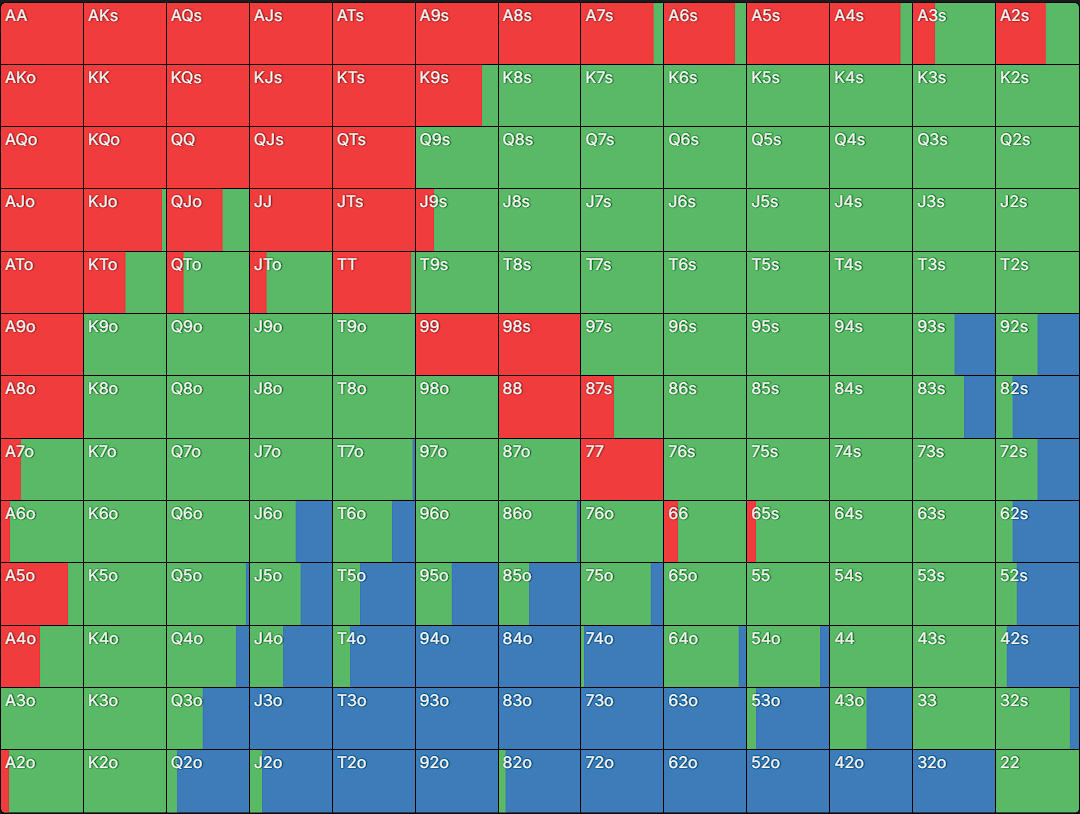}
    \caption{BB vs SB Open - Claude Opus 4.6}
    \label{fig:bb-vs-sb-claude}
\end{figure}
\begin{figure}
    \centering
    \includegraphics[width=0.7\linewidth]{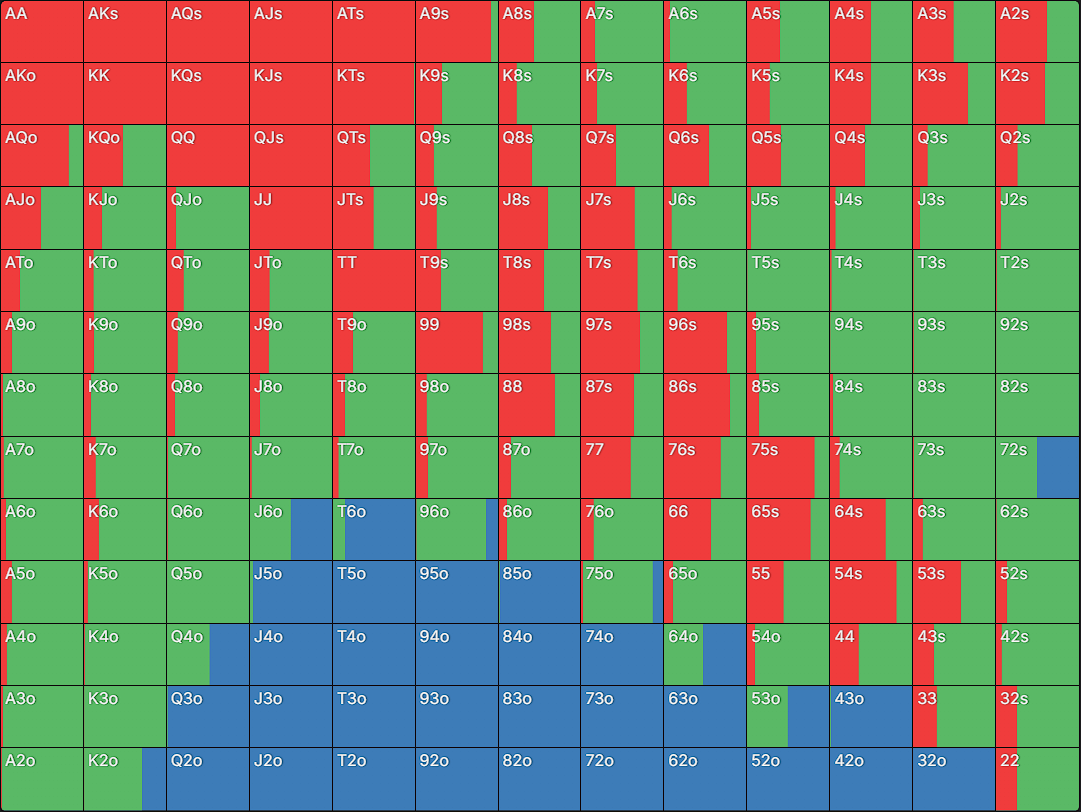}
    \caption{BB vs SB Open - GTO Wizard AI}
    \label{fig:bb-vs-sb-gtow}
\end{figure}

Crucially, the performance gaps are not only strategic but also representational. The 2\% suit-mismatch error observed by GPT-5.3, while small, indicates that even state-of-the-art LLMs struggle with the precise state tracking required for sound strategy. In a game like HUNL, where a single misunderstood blocker can change the expected value of a decision, these minor hallucinations compound, preventing even the best generalist models from approaching the performance of GTO Wizard AI. 

While there is a clear trend of improvement in the poker-playing ability of recent reasoning models, these models are likely highly exploitable due to their deterministic decision-making. In fact, LLMs do not seem to consider the probability distribution of cards they and their opponents could hold (known as beliefs or ranges in poker) when thinking about how they should act. Instead, they seem to reason on a per-hand basis to maximize their expected value, which makes them unbalanced and likely highly exploitable. This becomes evident when comparing GTO Wizard AI’s ranges to the LLM ranges as in the above table.
LLM ranges include significantly less strategic mixing, which makes their strategies more predictable and causes their actions to correlate too strongly with specific hands or public actions. This predictability leaks information to an opponent, who can then condition their responses to exploit those imbalances.

\subsubsection{Beyond win rates vs GTO Wizard AI}
An important thing to note is that this benchmark evaluates agents against our best (static) approximation of a Nash Equilibrium. In two-player, zero-sum games of imperfect information, it can be helpful to classify mistakes into one of two types: Expected Value (EV) mistakes and frequency mistakes. EV mistakes refer to when a player chooses a zero-frequency action that loses EV against a perfect opponent. Frequency mistakes (or mixing mistakes) refer to mistakes where a player is taking an action that should be taken at equilibrium, but does so with the wrong frequency. Both types of mistakes create opportunities for exploitation by strong opponents. As an example consider the standard game rock-paper-scissors (RPS).
In RPS, there are no EV mistakes because all actions have an EV of 0 against an opponent playing a Nash Equilibrium, but any strategy that is not part of a Nash Equilibrium (i.e. not uniform random) would constitute a frequency mistake, because it opens up the player to be exploited if their opponent is cognizant of that fact. In complex games like poker, it turns out to be quite difficult to completely avoid EV mistakes without a strong understanding of the game dynamics. Since GTO Wizard AI currently doesn't model and adjust to specific opponents, this benchmark only evaluates EV mistakes. 

As demonstrated by our results, current LLMs struggle with EV mistakes, but an important avenue for future work incorporates opponent modeling to our benchmark so that it can also evaluate frequency mistakes.
Such an improvement would enable the benchmark to evaluate agents on their exploitability, rather than head-to-head performance with GTO Wizard AI, which plays a strategy with close to zero exploitability itself.

\section{Conclusion}
In this work, we introduced GTO Wizard Benchmark, a rigorous framework for evaluating the capabilities of agents in poker. By leveraging AIVAT, a variance reduction technique, we drastically reduced the number of hands needed to obtain statistically significant results compared to current evaluation methods.

Our extensive evaluation of leading reasoning-based LLMs reveals a rapidly evolving landscape. The progression from GPT-4’s loss rate of -136.2 bb/100 to GPT-5.3’s -16 bb/100 marks a steep improvement in reasoning capabilities over a span of less than three years. This leap suggests that generalist agents under zero-shot settings are beginning to internalize the complex subtleties of poker strategy, such as equity, blockers, and long-term planning that were previously the exclusive domain of specialized solvers.

However, the gap between superhuman and current reasoning models remains significant. While models are progressing toward the level of human professional play, they are far from the superhuman baseline set by GTO Wizard AI. The observed state-tracking errors and hallucinations highlight that, while reasoning models are becoming more sophisticated, they still lack some of the fundamentals required for superhuman poker play. By offering a standardized task with clear room for progress, the benchmark allows researchers to rigorously compare models and methods and demonstrate quantifiable evidence of progress.

Ultimately, poker remains a challenging benchmark for multi-agent reasoning under partial observability. It demands long-term planning, navigating uncertainty, and a great understanding of the impact of the decisions made by the agent and their opponents on expected value. GTO Wizard Benchmark provides the necessary tooling to measure this progress. It will serve as a valuable asset for accelerating the development of smarter agents and help identify new methods that lead to robust reasoning under partial observability.

\bibliographystyle{plain}
\bibliography{references}

\section{Appendix}
\subsection{Heads-Up No-Limit Hold'em}
Heads-Up No-Limit Texas Hold'em (HUNL) is a two-player, zero-sum game of imperfect information. The particular variant of the rules described here is standardized for academic research and computer poker competitions. In this formulation, each game, or "hand," is treated as an independent event, with player chip stacks resetting to a starting value of $20,000$ before each hand begins

The game commences with each player receiving two private cards, known as "hole cards," from a standard 52-card deck. A forced betting structure is initiated through "blinds." One player, designated the small blind (SB), posts a mandatory bet of $50$, while the opponent, the big blind (BB), posts $100$. These positions alternate between the two players every hand. The first betting round, termed the "preflop," begins with the SB acting first. Following the conclusion of the preflop betting round, three community cards are dealt face-up. This is known as the "flop." These cards are available to both players. A second betting round ensues, initiated by the player in the SB position. Subsequently, a fourth (the "turn") and fifth (the "river") community card are dealt, each followed by its own betting round. The betting order for all postflop rounds remains consistent, with the SB player acting first. 

During a betting round, a player may choose one of five actions:
\begin{itemize}
    \item Fold: The player forfeits their cards and any claim to the pot, immediately ending the hand and awarding the pot to the opponent.
    \item Check: The player continues playing in the betting round without wagering any chips. This action can only be taken if no previous bets have been made during the betting round.
    \item Bet: The player places a wager during the betting round.
    \item Call: The player matches the amount wagered by the opponent in the current round.
    \item Raise: The player bets more chips than the previous bet made by the opponent in the current betting round.
\end{itemize}
A betting round concludes when both players have contributed equal amounts to the pot or one player folds. A single iteration of the full process described above is referred to as a “hand”. If the hand proceeds through all four betting rounds without a player folding, it culminates in a "showdown." At this stage, each player constructs the best possible five-card poker hand using any combination of their two private hole cards and the five community cards. The player with the hand of superior rank, according to the standard poker hand hierarchy, wins the entire pot. In the case of identical hand rankings, the pot is divided equally between the players.

\subsection{Prompt}
We use the following prompt for agent evaluation:

\texttt{You are an expert poker player playing heads-up No-Limit Texas Hold'em.
Your sole objective is to maximize your expected value against GTO Wizard, an AI solver trying to approximate a Nash Equilibrium. 
Game State Information:
You will receive comprehensive game state data including:
\begin{itemize}
\item hand\_id: Unique identifier for the current hand
\item game: Game configuration (blinds, starting stack, format)
\item game\_state: Current decision point with the following fields:
\begin{itemize}
    \item street: Current betting round ("preflop", "flop", "turn", or "river")
    \item board\_cards: Community cards (empty string on preflop, e.g., "AhKd3c" on flop)
    \item common\_pot: Chips in the middle from completed betting rounds
    \item total\_pot: Total chips committed (common\_pot + current round bets)
    \item players: List of two players with:
    \begin{itemize}
        \item name: Player identifier
        \item position: "BB" (Big Blind) or "SB" (Small Blind)
        \item stack: Remaining chips
        \item hole\_cards: Your private cards (visible only for you, null for opponent)
        \item legal\_actions: Base actions you can take (subset of ["f", "c", "k", "b"])
        \item raise\_range: If "b" is legal, the min/max chip amounts for betting/raising
        \item action\_history: Complete sequence of actions in this hand (see format below)
        \item is\_hand\_over: Whether the hand has concluded
        \item has\_gto\_wizard\_folded: Whether your opponent has folded
        \item winnings: Your chips won/lost (relevant when hand is over)
    \end{itemize}
\end{itemize}
\end{itemize}
Response Format:
You must respond with a JSON object containing:
\begin{itemize}
    \item reasoning: A brief 1-2 sentence explanation of your strategic thinking for this play
    \item action: One of the legal\_actions ("f", "k", "c", or "b")
    \item amount: Required only for "b" action, must be within raise\_range.min and raise\_range.max
\end{itemize}
Action History Format:
The action\_history shows all actions sequentially from the start of the hand:
\begin{itemize}
    \item 'f': Player folded
    \item 'c': Player called the current bet
    \item 'k': Player checked
    \item 'bX': Player bet/raised to X total chips in this betting round
    \item '\_': Betting round ended (separator between preflop/flop/turn/river)
\end{itemize}
Example action\_history: ['b200', 'b800', 'c', '\_', 'k', 'b1600', 'c', '\_', 'k', 'k', '\_', 'k', 'k']
\begin{itemize}
    \item Preflop: SB bet 200, BB raised to 800, SB called
    \item Flop: BB checked, SB bet 1600, BB called
    \item Turn: Both players checked
    \item River: Both players checked
\end{itemize}
Important Notes:
\begin{itemize}
    \item You are always one of the two players; never GTO Wizard.
\end{itemize}
}
\subsection{Evaluation Metrics}
Below is the full list of evaluation metrics returned by GTO Wizard Benchmark's API:
\begin{itemize}
    \item Total number of hands played.
    \item AIVAT Score: the number of GTO Wizard AI's chips the agent was expected to win, adjusted for luck, expressed in big blinds per 100 hands.
    \item bb/100: the number of big blind the agent won or lost on average for every 100 hands played against GTO Wizard AI.
    \item Chips: the number of raw chips won or lost by the agent.
    \item All Hands Chips: the number of GTO Wizard’s chips the agent won, playing with their full range, rather than an individual hand, expressed in big blinds per 100 hands.
    \item Chance Correction: an estimate of how lucky the agent's cards and board cards were, expressed in big blinds per 100 hands.
    \item Action Correction: an estimate of how lucky the GTO Wizard's actions were for the agent, expressed in big blinds per 100 hands.
    \item AIVAT standard deviation: standard deviation of the agent's cumulative AIVAT score across all hands, expressed in big blinds per 100 hands.
\end{itemize}

\subsection{Detailed agents’ performance against GTO Wizard AI}
We report the detailed evaluation of agents in Table \ref{tab:agentevaluation}

\begin{table}[ht]
    \centering
    \small
    \begin{tabularx}{\textwidth}{>{\raggedright\arraybackslash}p{0.28\textwidth} *{6}{>{\centering\arraybackslash}X}}
        \toprule
        \textbf{Agent} & \textbf{AIVAT Score} & \textbf{AIVAT Stddev} & \textbf{Chips Won} & \textbf{All Hands Chips} & \textbf{Chance Correction} & \textbf{Action Correction} \\
        \midrule
        GPT-5.3 Extra High reasoning & -16.04 & 3.04 & -26.34 & -25.98 & -11.36 & 1.42 \\
        GPT-5.4 Extra High reasoning & -17.84 & 3.71 & 29.05 & 5.52 & 9.59 & 13.77 \\
        GPT-5.3 High reasoning & -18.17 & 3.86 & -22.23 & -26.23 & -1.49 & -6.68 \\
        Claude Opus 4.6 & -20.35 & 4.39 & 20.94 & 5.62 & 7.20 & 18.83 \\
        Claude Opus 4.5 & -22.26 & 5.13 & -12.45 & -18.11 & 6.44 & -2.29 \\
        Gemini 3.1 Pro Preview & -30.83 & 2.30 & -30.38 & -31.08 & 0.65 & -0.69 \\
        Gemini 3 Pro & -30.06 & 3.85 & -44.72 & -25.40 & -4.41 & 9.07 \\
        Gemini 2.5 Pro & -39.19 & 4.91 & -24.66 & -46.18 & -0.67 & -6.32 \\
        Kimi K2.5 & -41.43 & 6.35 & -53.24 & -54.34 & -15.58 & 2.67 \\
        GPT-5.4 (No Reasoning) & -57.50 & 6.87 & -54.93 & -55.63 & -8.92 & 10.79 \\
        Always Fold & -64.56 & 1.71 & -63.12 & -63.11 & 1.37 & 0.08 \\
        Grok 4 High reasoning & -59.98 & 7.69 & -85.74 & -50.31 & 10.08 & -0.55 \\
        GPT-4o & -60.90 & 7.63 & -58.69 & -43.24 & 8.53 & 9.06 \\
        GPT-5.4 Mini (No reasoning) & -107.87 & 9.08 & -159.12 & -157.80 & -28.07 & -21.86 \\
        GPT-4 & -136.17 & 13.08 & -173.11 & -148.17 & -13.33 & 1.22 \\
        GPT-5.4 Nano & -189.73 & 15.27 & -190.22 & -187.76 & -17.60 & 19.57 \\
        Check call & -241.01 & 13.39 & -269.33 & -229.66 & -0.48 & 11.89 \\
        Random Uniform & -284.85 & 15.09 & -301.71 & -311.99 & -5.21 & -21.93 \\
        All-in & -380.57 & 2.18 & -332.42 & -393.81 & 2.51 & -15.74 \\
        \bottomrule
    \end{tabularx}
    \caption{Detailed agent evaluation. All the metric values are in big blinds per hundred hands (bb/100). Models are ranked according to the lower bound of their 95\% confidence interval AIVAT score. }
    \label{tab:agentevaluation}
\end{table}

\subsection{Suit Mismatch Examples}
\textbf{Example 1}
\texttt{
\\
Model: GPT-5.3 Extra High reasoning
\\
Hand: Qd9s
\\
Reasoning: Q9s is a clear continue versus a 2.25x SB open: we only need about 28\% equity to call and this hand has strong playability against a wide opening range. It can mix into some 3-bets, but flatting is the highest-EV baseline here at 200BB deep.}

\textbf{Example 2}
\texttt{
\\
Model: GPT-5.3 Extra High reasoning
\\
Hand: Ah6s
\\
Reasoning: A6s is a clear profitable small-blind open in heads-up play, benefiting from blocker value and strong suited playability in position. I’ll apply pressure with a standard raise size rather than limp, which captures more immediate fold equity while keeping initiative.}

\textbf{Example 3}
\texttt{
\\
Model: Claude Opus 4.5
\\
Hand: KhJs
\\
Reasoning: KJs is a strong suited broadway hand that plays well in heads-up. From the SB, I should raise to build the pot with this premium-ish holding. A standard 2.5-3x raise is appropriate here.}

\textbf{Example 4}
\texttt{
\\
Model: Claude Opus 4.5
\\
Hand: Kh9s
\\
Street: River
\\
Position: Big Blind
\\
Reasoning: I have middle pair with K9s on a board of 8c5s7hQs9d. The action has been very passive - SB limped preflop and both players checked flop, turn, and now it's checked to me on the river. My 9 is likely the best hand given the passivity, but betting for thin value risks getting check-raised or called by better hands like Qx, two pair, or straights (J-T, 6-4). A small value bet could get called by worse pairs like 8x or 7x, but checking back is safer to realize my showdown value.}

\textbf{Example 5}
\texttt{
\\
Model: Claude Opus 4.5
\\
Hand: QcTs
\\
Reasoning: QTs is a strong suited broadway hand that plays well postflop. From the SB, this is a clear raise for value and position. Standard open to 2.5x.
}
\subsection{Card Mismatch Examples}
\textbf{Example 1}
\texttt{
\\
Model: GPT-4
\\
Hand: 8c7s
\\
Reasoning: 86s is a decent and playable hand, I will call to see the flop.
}

\textbf{Example 2}
\texttt{
\\
Model: GPT-4
\\
Hand: Ks4h
\\
Board: 2d6s5s5c5h
\\
The board is showing a five-of-a-kind, meaning both players have an equivalent five-card hand of 55555. So, no matter the hole cards, the round will likely end in a draw, unless the opponent folds. Let's hope for a potential overbet fold.
}

\textbf{Example 3}
\texttt{
\\
Model: Claude Opus 4.5
\\
Hand: 9d7s
\\
Board: 8d4c3sKs5d
\\
Reasoning: I have 9d7s on a Kd4c3sKs board - complete air with no pair, no draw, and only a gutshot to a 6 for a straight. The board has paired Kings and two spades. After both players checked flop and opponent checked turn, this could be a good spot to bluff representing a King or strong hand, but my hand has very little equity if called. With no showdown value and facing a turned paired board, I should check and try to see a free river where I might improve or can reassess bluffing opportunities.
}

\textbf{Example 4}
\texttt{
\\
Model: Gemini 3 Pro
\\
Hand: Kc2c
\\
Board cards: 5sQsJc8dJs
\\
Reasoning: On a dynamic QsQsJc board, K2s has a gutshot to the nuts and backdoor flush equity, but weak showdown value and cannot withstand a check-raise. Checking back allows us to realize equity for free and keeps the pot manageable with a marginal hand.
}
\end{document}